\documentclass[sigconf]{acmart}
\AtBeginDocument{%
  }

\setcopyright{acmlicensed}
\copyrightyear{2026}
\acmYear{2026}
\acmDOI{XXXXXXX.XXXXXXX}
\acmConference[ICSE '26]{48th International Conference on Software Engineering}{April 2026}{Rio De Janeiro, Brazil}
\acmISBN{978-1-4503-XXXX-X/18/06}

\usepackage{amsmath,amsfonts,amsthm}
\usepackage{algorithm}
\usepackage{algpseudocode}
\usepackage{graphicx}
\usepackage{textcomp}
\usepackage{xcolor}
\usepackage{balance}
\usepackage{xspace}
\usepackage{enumitem}
\usepackage{cleveref}
\usepackage[many]{tcolorbox}
\usepackage{subfigure}
\usepackage{booktabs}
\usepackage[normalem]{ulem}
\usepackage{siunitx}
\usepackage{colortbl}
\usepackage{multirow}
\usepackage{array}
\usepackage{threeparttable}
\usepackage{pifont}
\usepackage{makecell} %
\usepackage{tabularx} %
\usepackage{listings}
\usepackage{wrapfig}
\usepackage{graphicx}                  %
\usepackage{bm}                        %
\usepackage{newunicodechar}            %

\usepackage{tikz}
\usetikzlibrary{positioning, arrows}

\newunicodechar{≡}{\equiv}
\newunicodechar{≥}{\geq}
\newunicodechar{≤}{\leq}

\newcommand{\scenethesistool}{\textsc{Scenethesis}\xspace}
\newcommand{\scenedsl}{\textsc{ScenethesisLang}\xspace}

\newcommand{\todoaftersub}[1]{\mytodomagenta{}}

\newcommand{\mytodomagenta}[1]{\mytodocomment{magenta}{{\sf}~#1}}

\newcommand{\mytodocomment}[2]{\textcolor{#1}{#2}}

\definecolor{applegreen}{rgb}{0.55, 0.80, 0.4}
\definecolor{lightblue}{RGB}{51, 153, 255}
\definecolor{deepyellow}{RGB}{214, 195, 49}

\let\diff\undefined %

\ifdefined\diff
\newcommand{\textred}[1]{\textcolor{red}{#1}}
\newcommand{\textblue}[1]{\textcolor{blue}{#1}}

\newcommand{\textyellow}[1]{\textcolor{deepyellow}{#1}}
\else
\newcommand{\textred}[1]{#1}
\newcommand{\textblue}[1]{#1}

\newcommand{\textyellow}[1]{#1}
\fi

\lstdefinestyle{mystyle}{
    commentstyle=\color{applegreen},
    keywordstyle=\color{blue},
    stringstyle=\color{purple},
    basicstyle=\ttfamily\scriptsize,       %
    breakatwhitespace=false,
    breaklines=true,
    captionpos=b,
    keepspaces=true,
    numbers=left,
    numbersep=5pt,
    showspaces=false,
    showstringspaces=false,
    showtabs=false,
    tabsize=2,
    frame=single,                     %
    rulecolor=\color{black}           %
}
\copyrightyear{2026}
\acmYear{2026}
\setcopyright{cc}
\setcctype{by}
\acmConference[ICSE '26]{2026 IEEE/ACM 48th International Conference on Software Engineering}{April 12--18, 2026}{Rio de Janeiro, Brazil}
\acmBooktitle{2026 IEEE/ACM 48th International Conference on Software Engineering (ICSE '26), April 12--18, 2026, Rio de Janeiro, Brazil}
\acmPrice{}
\acmDOI{10.1145/3744916.3787812}
\acmISBN{979-8-4007-2025-3/2026/04}
\begin{document}

\title{3D Software Synthesis Driven by Constraint-Expressive Intermediate Representation}
\author{Shuqing Li}
\orcid{0000-0001-6323-1402}
\affiliation{%
  \institution{The Chinese University of Hong Kong}
  \city{Hong Kong}
  \country{China}
}
\email{sqli21@cse.cuhk.edu.hk}

\author{Anson Y. Lam}
\orcid{0009-0006-3976-4675}
\affiliation{%
  \institution{The Chinese University of Hong Kong}
  \city{Hong Kong}
  \country{China}
}
\email{yflam1@cse.cuhk.edu.hk}

\author{Yun Peng}
\authornote{Yun Peng and Wenxuan Wang are corresponding authors.}
\orcid{0000-0003-1936-5598}
\affiliation{%
  \institution{The Chinese University of Hong Kong}
  \city{Hong Kong}
  \country{China}
}
\email{ypeng@cse.cuhk.edu.hk}

\author{Wenxuan Wang$^{*}$}
\orcid{0000-0002-9803-8204}
\affiliation{%
  \institution{Renmin University of China}
  \city{Beijing}
  \country{China}
}
\email{wangwenxuan@ruc.edu.cn}

\author{Michael R. Lyu}
\orcid{0000-0002-3666-5798}
\affiliation{%
  \institution{The Chinese University of Hong Kong}
  \city{Hong Kong}
  \country{China}
}
\email{lyu@cse.cuhk.edu.hk}

\renewcommand{\shortauthors}{Li et al.}

\begin{abstract}
Graphical user interface (UI) software has undergone a fundamental transformation from traditional 2D
interfaces to spatial 3D environments.
While existing work has made remarkable success in
2D software generation,
3D software generation still remains underexplored.
Current methods for 3D software generation usually generate 3D environment as a whole and cannot modify
specific elements.
Furthermore, these methods struggle to handle the complex spatial and semantic constraints inherent in the real world.

To address these challenges, we present \scenethesistool, a novel requirement-sensitive 3D software synthesis approach that maintains formal traceability between user specifications and generated 3D software.
\scenethesistool is built upon \scenedsl, a domain-specific language that serves as a granular constraint-aware intermediate representation to bridge natural language requirements and
3D software. 
\textyellow{
It serves both as a comprehensive scene description language enabling fine-grained modification of 3D software elements and as a formal constraint-expressive specification language capable of expressing complex spatial constraints.
}
By decomposing 3D software synthesis into stages,
\scenethesistool enables independent verification, targeted modification, and systematic constraint satisfaction.
Our evaluation 
demonstrates that \scenethesistool accurately captures over 80\% of user requirements and satisfies more than 90\% of hard constraints while handling over 100 constraints simultaneously. Furthermore, \scenethesistool achieves a 42.8\% improvement in BLIP-2 visual evaluation scores compared to the state-of-the-art method, establishing its effectiveness in generating high-quality 3D software that faithfully adheres to complex user requirements.
\textyellow{Source code and supplemental materials are publicly available at \url{https://sites.google.com/view/3d-software-synthesis}.}
\end{abstract}

\maketitle

\vspace{-1.5em}
\section{Introduction}

Graphical user interface (UI) software has been a cornerstone of computing since the introduction of the Xerox Alto in 1973~\cite{thacker1979alto}.
\textred{Over the past decade, the software engineering (SE) community has developed mature, well-established ecosystems and techniques for automated 2D UI generation~\cite{uicopilot2024, design2code2024, webcode2m2024, dcgen2024, layoutcoder2024, xiao2024interaction2code}, ranging from template-driven synthesis and constraint-based layout algorithms to model-based approaches and even methods powered by cutting-edge Large Language Models (LLMs).} Driven by advances in graphics hardware and
engines such as Unity, 3D software has experienced explosive growth since the early 2000s.
The global 3D software market reached more than 32 billion in 2024~\cite{3d-market}, \textred{spanning many SE-related domains such as robotics~\cite{yu2024practical,freitas2024rfnit,humeniuk2024simulation}, training platforms for autonomous (aerial) vehicles~\cite{sorokin2024opensbt,khan2024simulation,steininger2025automatically,song2024an}, 3D games, virtual production systems, modeling and design applications,
and Extended Reality (VR/AR/MR) applications~\cite{karre2025vreqst,nusrat2021how,guo2024an,wang2024vrguide,rafi2023predart,mehra2020xrase}}. Despite the rapid growth of 3D software, the automated synthesis of 3D software still remains underexplored.

\textyellow{
The established methods for 2D UI generation cannot be directly applied to 3D software synthesis due to fundamental differences in spatial complexity, physical constraints, and interaction paradigms.
}
Recent end-to-end text-to-3D generation approaches propose directly generating complete 3D software from natural language (NL) based on neural synthesis~\cite{hollein2023text2room,li2024dreamscene}, procedural modeling~\cite{procthor}, or constraint-based methods~\cite{yang2024holodeck}. They typically regard 3D software generation as a monolithic vision problem rather than a structured software synthesis task. 
However, high-quality 3D software should not only be visually compelling but also functionally correct, physically plausible, and programmatically testable. These approaches lack fine-grained intermediate representations (IRs) that bridge the semantic gap between high-level requirements and low-level
implementations.
Some work~\cite{yang2024holodeck} pioneered the use of intuitive IR, such as scene graphs, to capture users' requirements.
While intuitive, scene graphs restrict object classes to predefined categories and relationships to a few discrete types (e.g., \textit{left} and \textit{top}), which fundamentally limits their expressiveness. Moreover, the assumption of at most one relation between two objects makes it impossible to express complex spatial constraints for real-world applications. 

Specifically, current approaches to 3D software synthesis face the following challenges:
\textbf{Challenge 1 (C1): Lack of Compositional Control and Post-Generation Maintainability.} 
Current \textblue{3D software synthesis} methods generate 3D software as a whole and do not support modification of specific elements in the software. The lack of controllability over specific elements makes it quite challenging to meet precise specifications, as \textblue{these} methods have to regenerate the entire software in each iteration to fix even minor errors.
This is a fundamental violation of SE principles of predictability and control.
Furthermore, when specifications evolve or bugs are discovered in deployed 3D software, developers cannot perform targeted fixes or incremental updates.
The absence of expressive IRs between requirements and final 3D software fundamentally prevents developers from tracing the rationale behind specific design decisions and maintaining version control at the component level.
\textbf{Challenge 2 (C2): Inability to Handle Complex Constraints.} 
Real-world 3D software systems require satisfying diverse spatial, semantic, and physical constraints. For instance, a robot testing environment might require ``\textit{all emergency equipment must be accessible within 2 meters from any workstation.}''
Current methods cannot reliably encode or verify such domain-specific requirements. Structure-based approaches like InstructScene~\cite{lin2024instructscene} employ scene graphs to illustrate the complex constraints, but they suffer from severe expressiveness limitations.
\textyellow{
Scene graphs incorporate only simple and fixed spatial relationship categories, such as ``left'' and ``above'', to describe the constraints between objects, so they can hardly capture the complicated continuous spatial relationships required by user constraints.
}

To address these challenges, we present \scenethesistool, a novel UI code synthesis system for 3D software environments. \textred{In stark contrast to existing monolithic, black-box approaches, our method introduces a principled, requirements-driven process rooted in core SE principles.} \textred{At its heart lies \scenedsl, a novel modeling domain-specific language (DSL) that makes a primary contribution to requirements engineering by enabling robust translation of high-level, ambiguous NL to precise, verifiable specifications.} \textred{The entire synthesis pipeline is structured as a model-driven engineering (MDE) process~\cite{liu2025tbfv4j,chen2025gpass,sanchez2025qedcartographer,ma2024specgen,zheng2024testing,cordy2023towards,albataineh2024a} where \scenedsl serves as the central formal model.} This MDE approach fundamentally reimagines 3D software synthesis through an SE lens, \textred{overcoming the opaqueness of end-to-end models and the rigidity of scene-graph-based methods by} \textred{decomposing the complex problem of 3D software synthesis into four distinct, verifiable, and interconnected stages that collectively maximize correctness and ensure tractability}:

\textbf{Requirement Formalization}: \scenethesistool translates NL requirements into formal \scenedsl specifications, establishing unambiguous semantics for all software assets (i.e., objects in 3D environments) and spatial relationships.
\textyellow{
\scenedsl also encodes implicit physical laws that are often overlooked by users in the requirements but must be followed to make the generated software both physically plausible and functionally correct.
}
\textbf{Asset Synthesis}: \scenethesistool transforms object declarations from \scenedsl specifications into concrete 3D models through a hybrid strategy, which balances the retrieval of existing models from curated databases and the text-to-3D generation of new models. This strategy ensures both quality and coverage.
\textbf{Spatial Constraint Solving}: By formulating object placement as a constraint satisfaction problem over continuous 3D space, we design a novel Rubik Spatial Constraint Solver that employs an iterative refinement approach inspired by Rubik's Cube solving, where local adjustments propagate to achieve global constraint satisfaction. This method \textred{maximizes constraint satisfaction rate but at the same time remains computationally tractable, even for complex scenarios with hundreds of spatial constraints}.
\textbf{Software Synthesis}: The final stage combines solved object layouts with acquired 3D models to produce executable Unity-compatible software artifacts.
\textred{After a software is synthesized, if a user wants to edit or update the software, \textbf{targeted modification} can be achieved by either using \scenethesistool to automatically modify the scene or manually modifying the human-readable IR.}

This modular, inspectable generation pipeline provides transparency and control
, allowing developers to inspect IRs and modify specific components without full regeneration\textred{, a limitation of monolithic systems that forces costly and unpredictable regeneration for any minor change}. \scenedsl enables developers to express arbitrary spatial, physical, and semantic constraints using a rich algebra of operations and predicates. It also moves beyond the categorical limitations of scene graphs
to support
multiple simultaneous relationships and complex logical compositions.
To evaluate \scenethesistool, we construct a dataset consisting of 50 highly comprehensive user queries,
spanning a diverse spectrum of room types. Evaluation results show that \scenethesistool can accurately capture more than 80\% of user requirements even when the threshold is relatively high, and that it can satisfy over 90\% of hard constraints while handling more than 100 constraints. In terms of visual scores, \scenethesistool outperforms all baselines (end-to-end LLM and Holodeck~\cite{yang2024holodeck}) across all metrics under different LLM backbones, and even achieves a BLIP-2~\cite{li2023blip} evaluation score that is 42.8\% higher than current state-of-the-art, Holodeck~\cite{yang2024holodeck}.

The primary contributions of this work are as follows:
\begin{itemize}[leftmargin=*]
\item A formal DSL for 3D software that unifies spatial constraint specification with scene description, providing both expressiveness and verifiability for 3D software generation.
\item A
four-stage synthesis pipeline that decomposes 3D software generation into requirement formalization, asset synthesis, spatial constraint solving, and software synthesis, with each stage being independently verifiable and modular. \textred{Targeted modification can be done automatically or manually for post-editing.}
\item A novel iterative constraint-solving algorithm that avoids the
complexity of traditional approaches through local-to-global refinement, achieving practical scalability for complex 3D software.
\end{itemize}

\vspace{-1em}
\section{Preliminaries}
\textbf{3D Software.} 
3D software systems represent virtual environments through a hierarchical structure~\cite{foley2013computer}. A \emph{3D model} consists of three components: (1) \emph{Geometry}: meshes comprising vertices, edges, and faces that define surface structure; (2) \emph{Appearance}: materials specifying textures, colors, and shaders for visual rendering; (3) \emph{Spatial Properties}: transforms encoding position $(x,y,z)$, rotation $(r_x,r_y,r_z)$, and scale $(s_x,s_y,s_z)$ in 3D space.
A \emph{scene} comprises multiple models in a shared coordinate system. We adopt a left-handed system where (1) $x$, $y$, and $z$ axes represent width, height, and depth respectively, (2) the front of an unrotated object should be facing the positive $z$-axis, and (3) the order of rotation is $x \rightarrow z \rightarrow y$. Unity~\cite{unity}, our target platform, organizes content through GameObjects containing components (mesh renderers, colliders, scripts) that define 3D software entities.
\textred{Listing~\ref{code:unity} shows a code snippet (written in C\#) for generating a Unity scene.}

\textbf{Spatial Constraints.} 
Professional 3D software must satisfy three constraint categories:
\emph{Geometric constraints} specify spatial relationships (e.g., "A is 2m from B": $||pos_A - pos_B||_2 = 2.0$).
\emph{Physical constraints} ensure plausibility through collision avoidance and gravity support.
\emph{Semantic constraints} encode domain rules (e.g., "emergency exits must be accessible").

\section{Approach: \scenethesistool}
\label{sec:methodology}

 \begin{figure*}[t!]
 \vspace{-1em}
 	\centering 
 	\includegraphics[width=0.9\textwidth]{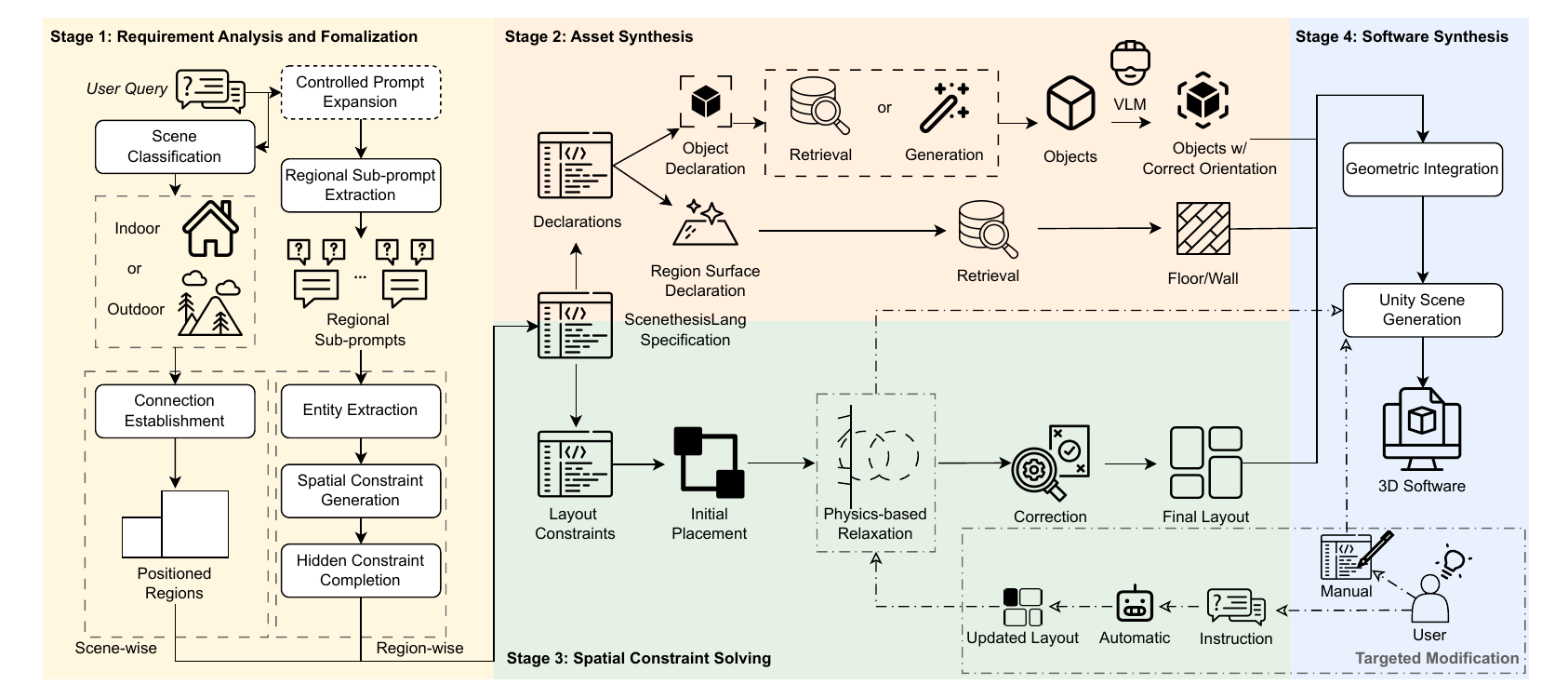} 
        \vspace{-2em}
 	\caption{\textred{Overview of \scenethesistool}}
 	\label{fig:scenethesis-overview}
       \vspace{-1.5em} 
 \end{figure*}

Throughout this framework, \scenedsl serves as the \linebreak[4]constraint-expressive IR, providing formal semantics for spatial relationships while making implicit physical laws (e.g., gravity, collision avoidance) explicit. Figure~\ref{fig:scenethesis-overview} shows the overview of \scenethesistool.

\begin{figure}[t!]
\scriptsize
\vspace{-1em}
\begin{minipage}{0.95\linewidth}
\begin{align}
program \in \mathit{Programs} ::=&\ stmt; \ | \ stmt;\ program\\
stmt \in \mathit{Statements} ::=& \ \mathit{decl} \ | \ \mathit{const} \ | \ assign \\
decl \in \mathit{Declarations} ::=& \ \textbf{object}\;id\ |\ \textbf{region}\;id \ | \ \tau \ id \\
const \in \mathit{Constraints} ::=&\ \textbf{assert}\;\phi \
             |\  \textbf{allowCollide}(id,id) \
             |\  \textbf{allowOutside}(id) \\
assign \in \mathit{Assignments} ::=&\ id.\alpha \leftarrow e \ | \ id.\beta \leftarrow e \ | \ id \leftarrow e\\
\alpha \in \mathit{Object\ Properties} ::= &\ \textbf{color} \mid \textbf{material} \mid \textbf{features}\\
\beta \in  \mathit{Region\ Properties} ::=&\ \textbf{pos} \mid \textbf{rot} \mid \textbf{scale} \mid  \\
\phi \in \mathit{Assertions}::=& \ e\,\bowtie\,e
     \mid \textbf{inside}(id,id)
     \mid \phi \land \phi
     \mid \phi \lor \phi
     \mid \lnot\phi \\
\tau \in \mathit{Types}::=&~\textbf{Number}\mid\textbf{Degree}\mid\textbf{Bool}\mid\textbf{Vector3} \\
&\mid \textbf{Rotation}\mid\textbf{Color}\mid\textbf{Material} \\
e \in  \mathit{Expressions} ::=&\ n \mid id \mid s
     \mid e\,\odot\,e \\
     &\mid \textbf{rand}(e,\ e)
     \mid \textbf{vec3}(e,\ e,\ e)
     \mid \textbf{rot}(e,\ e,\ e) \\
     &\mid \textbf{dot}(e,\ e)
     \mid id.p \\
\bowtie \in \mathit{Compare\ Operators} ::=&~= \mid \neq \mid < \mid \le \mid > \mid \ge \\
\odot \in  \mathit{Arithmetic\ Operators} ::=&~+ \mid - \mid * \mid / \\
n \in \mathit{Numbers} ::= &\ \text{any number} \\
s \in \mathit{Strings} ::= &\ \text{any string} \\
id \in \mathit{Identifiers} ::=&\ \text{names of the objects, regions and variables}
\end{align}
\vspace{-3em}
\caption{Domain-specific language: \scenedsl}
\label{fig:scenethesisdsl}
\end{minipage}
\vspace{-3em}
\end{figure}

\subsection{Stage I: Requirement Formalization}

The first stage transforms ambiguous NL input into a precise, verifiable specification in \scenedsl. This formalization process serves two critical goals: establishing unambiguous semantics for all requirements and inferring hidden physical constraints.

\subsubsection{Natural Language Analysis and Contextualization}

Given a user query $Q$, we first perform semantic analysis to determine the scene context and extract structured information. We employ an LLM with few-shot prompting to classify the scene type (indoor vs. outdoor), which determines the applicable constraint templates and default assumptions. For instance, indoor scenes automatically inherit boundary constraints (objects must remain within walls) and require ceiling/floor specifications, while outdoor scenes assume unbounded horizontal space.

\scenethesistool then performs LLM-based \textbf{controlled prompt expansion} to enrich the description with contextual details, since user requirements usually contain hidden constraints. For example, the user requirement ``a modern conference room'' contains hidden requirements for the furniture arrangements, lighting conditions, and accessibility. The expansion is strictly constrained to preserve all explicit user requirements while adding plausible hidden constraints inferred from the user requirements. Formally, let $Q^{'}$ denote the expanded prompt where $Q^{'} = Q \cup \{c_1, c_2, \ldots, c_k\}$ such that each $c_i$ represents an inferred contextual constraint from $Q$. 
Next, sub-prompt $Q^{'}_i$ ($i \in \left\{1,\dots,r\right\}$ where $r$ is the number of regions) for region $i$ is generated via another LLM call by extracting sentences in $Q^{'}$ that are relevant to region $i$. In the following stages, scene-wise processing and generation use $Q^{'}$ while region-wise processing and generation use $Q^{'}_i$.

\subsubsection{DSL Specification Generation}

$Q^{'}$ and $Q^{'}_i$ are then translated into a formal \scenedsl program consisting of declarations, constraints and assignments. Figure~\ref{fig:scenethesisdsl} presents the specification for \scenedsl. With object declaration statements, \scenedsl is able to describe each object in the scenes separately for stronger controllability. With constraint statements, \scenedsl could describe the arbitrary spatial relationships between objects to facilitate complex constraint solving.

Scene-wise, \scenethesistool first establishes connections between regions based on semantic information stored in $Q^{'}$. If the scene is indoor, the category (type of door or window), description, and dimensions of each connection object are also generated. \scenethesistool then proceeds to region positioning in which an LLM is asked to generate the vertices of each region conditioned on the previously established pairs of connections. To further improve realism, unlike previous work~\cite{yang2024holodeck}, we want walls in indoor scenes to have some thickness $\eta$ instead of being just a piece of paper. To this end, we begin by shifting the vertices of each region horizontally (the direction and amount to shift is outputted by an LLM) so that the distance between every neighboring pair is exactly $2\eta$ units. Then, when we are creating a mesh for each region in a later stage, we use Blender to extend the walls outward by $\eta$ units.

Region-wise, given $Q^{'}_i$, \scenethesistool conducts three steps to build the \scenedsl program:

\textbf{Step 1: Entity Extraction}. \scenethesistool first extracts all entities mentioned in $Q^{'}_i$ and creates an object declaration statement for each entity, i.e., $\textbf{entity}\ id$. Each entity in \scenedsl has three properties, \textit{color}, \textit{material} and \textit{features}, to describe the details of the entity. These entities can be divided into two types: region surface textures (floor and walls) and objects. Each object has two additional properties, namely, category and dimensions. \textred{In total, one region requires at least 10 statements (1 for $id$, 3 for floor, 3 for \texttt{pos}, 3 for \texttt{scale}), while one object requires at least 13 statements (1 for $id$, 3 for description, 3 for \texttt{pos}, 3 for \texttt{rot}, 3 for \texttt{scale}).}

\textbf{Step 2: Spatial Constraint Generation}. Given the extracted objects, \scenethesistool then captures the NL descriptions about the spatial relationships among the objects. For each captured spatial relationship, \scenethesistool creates a constraint statement in \scenedsl with the help of LLM to describe it. For instance, ``\textit{the lamp hangs above the table}'' becomes a constraint statement:
$\texttt{assert}~\textit{lamp}\texttt{.pos.y} > \textit{table}\texttt{.pos.y} + \textit{table}\texttt{.scale.y}$.

To reduce the chance of having redundant or contradictory subsets in the generated set of constraint statements (a set of constraints is redundant if keeping only one constraint in this set will not change the overall physical meaning of the scene, while a set of constraints is contradictory if satisfying one constraint in this set implies that other constraints in the same set can never be satisfied), we first pass the set to an LLM at most $\nu_r$ times to identify and remove redundant subsets, and then pass the resulting set to an LLM at most $\nu_c$ times to identify and remove contradictory subsets.

\textbf{Step 3: Hidden Constraint Completion}. Lastly, \scenethesistool adds physical realism constraints to make sure that the generated scene follows the sense of the physical world. For example, we add a constraint for all objects to ensure that they do not collide with each other unless allowed:
$\forall o_i, o_j \in \mathcal{O}, i \neq j \Rightarrow \neg\texttt{collides}(o_i, o_j) \vee \texttt{allowCollide}(o_i, o_j).$

Similarly, we also add gravity constraints to ensure proper support relationships, and boundary constraints to keep objects within designated regions unless explicitly overridden.

\textred{When generating constraints, we require the LLM to generate from 45 operations (e.g., \texttt{add(a: Number, b: Number) -> Number}) and 8 constraints (e.g., \texttt{mustEqualTo(a: Number, b: Number)}). \textred{Details can be further found on our website}. The operations come from numerous mathematical operators and functions defined in \scenedsl, as well as the ability of \scenedsl to get properties of entities. The constraints come from the comparison operators and \texttt{allow*()} statements defined in \scenedsl.}

\subsection{Stage II: Asset Synthesis}

Instead of generating the entire scene, \scenethesistool generates each object independently to ensure high controllability and facilitate easier fixes for small errors. The second stage processes object declarations from the \scenedsl specification to obtain concrete 3D models. This stage operates independently on each object, enabling parallel processing and modular replacement of acquisition strategies.

\textbf{Query Formulation.}
For an object, 
query is formulated as ``a 3D model of a <color> <category> made with <material> that is <features>''. 
For a region surface texture, 
query is formulated as ``a <color> floor/wall made of <material> that is <features>''. Note that in this paper, the surface texture of a region can only be retrieved.
\textbf{Hybrid Synthesis Strategy.}
To generate the object based on $q_o$, we employ a two-tier acquisition strategy that balances quality and coverage:
\textbf{(1) Retrieval-Based Acquisition}. Given query $q_o$, we first search a curated model database $\mathcal{D}$ using a composite similarity function:
$o^* = \arg\max_{o \in \mathcal{D}} score_{\text{ret}}(o, q_o)$,
where
$score_{\text{ret}}(o, q_o) = \frac{\lambda_v \cdot \textit{sim}_{\text{visual}}(o, q_o) + \lambda_t \cdot \textit{sim}_{\text{semantic}}(o, q_o)}{\lambda_v + \lambda_t}\;\text{,}$
$\textit{sim}_{\text{visual}}$ measures visual similarity (normalized to $\left[0, 1\right]$) using CLIP embeddings~\cite{radford2021learning} of rendered model views, and $\textit{sim}_{\text{semantic}}$ computes semantic similarity (normalized to $\left[0, 1\right]$) between textual descriptions using Sentence-BERT~\cite{reimers2019sentence}. The weights $\lambda_v$ and $\lambda_t$ are tuned empirically to balance visual fidelity and semantic accuracy.
\textbf{(2) Generative Acquisition}. If no suitable model is found in $\mathcal{D}$, i.e., $\max_{o \in \mathcal{D}} score_{\text{ret}}(o, q_o) < \tau$ for some threshold $\tau$, we invoke a text-to-3D generation technique.
Any acquired object is checked by a vision language model (VLM) to ensure that it is oriented canonically. Specifically, we \textblue{do so in the $x$-$z$-$y$ order. For each axis, we rotate the target object along that axis by $0^{\circ}$, $90^{\circ}$, $180^{\circ}$, and $270^{\circ}$, each time rendering the camera view. The rendered images are pasted on the top-left, top-right, bottom-left, and bottom-right corners of a larger image, respectively. The combined image is sent to the VLM for determining the rotation required along that axis to put the object in upright and front-facing orientation. In subsequent checks (i.e., along the $z$- and $y$-axes), the target object is rotated according to previous results.}

\subsection{Stage III: Spatial Layout Solving}

With the generated separate objects, the next step is to organize them properly in the scene.
The third stage constitutes the core innovation of our approach: formulating scene layout as a constraint satisfaction problem over continuous 3D space. \textred{This principled approach maximizes constraint satisfaction rate while remaining computationally tractable.}

 \begin{figure}[tbp]
 	\centering 
 	\includegraphics[width=0.4\columnwidth]{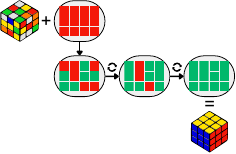} 
        \vspace{-1.5em}
 	\caption{Rubik Spatial Constraint Solver}
 	\label{fig:puzzling-constraint-solving}
    \vspace{-1em}  
 \end{figure}

\textred{As shown in Figure~\ref{fig:puzzling-constraint-solving},} our solver employs a novel iterative approach inspired by Rubik's Cube solving, where local adjustments propagate to achieve global constraint satisfaction. Algorithm~\ref{alg:constraint-solver} presents the complete procedure.

\begin{algorithm}[tbp]
\caption{Rubik Spatial Constraint Solver}
\scriptsize
\label{alg:constraint-solver}
\begin{algorithmic}[1]
\Require{Object set $\mathcal{O}$, constraint set $\mathcal{C}$, batch size $k$, max iterations $T$}
\Ensure{Valid layout $L^*$ satisfying all hard constraints in $\mathcal{C}$}
\State $L_0 \gets \textsc{InitialPlacement}(\mathcal{O})$
\State $L_0 \gets \textsc{PhysicsRelaxation}(L_0)$
\For{$t = 1$ to $T$}
    \State $\mathcal{U} \gets \{c \in \mathcal{C} : \neg\textsc{Satisfied}(c, L_{t-1})\}$
    \If{$|\mathcal{U}| = 0$}
        \State \Return $L_{t-1}$ 
    \EndIf
    \State $\mathcal{B} \gets \textsc{SelectBatch}(\mathcal{U}, k)$
    \State $L_t \gets \textsc{LLMSolve}(L_{t-1}, \mathcal{B}, \mathcal{C})$
    \State $L_t \gets \textsc{EnforceBounds}(L_t)$
\EndFor
\State \Return $\textsc{BestSolution}(L_0, L_1, \ldots, L_T)$
\end{algorithmic}
\end{algorithm}

The algorithm begins with \textsc{InitialPlacement} (line 1), which generates a baseline layout by considering all constraints at once. \textsc{PhysicsRelaxation} (line 2) applies basic collision resolution to create a physically stable starting configuration.

The core solving loop (lines 3$\sim$10) iteratively addresses unsatisfied constraints in batches. For each batch $\mathcal{B}$ of size $k$, we invoke \textsc{LLMSolve}, which leverages the spatial reasoning capabilities of LLMs to suggest object transformations. \textred{In particular, the LLM first} receives a structured description of the current layout, the violated constraints, and the complete constraint set. \textred{Then, utilizing its world knowledge, the LLM is able to propose} specific adjustments (translations, rotations) to resolve the violations. \textred{Though, since LLMs are still not the perfect constraint solvers, sometimes the adjustments proposed by the LLM might dissatisfy some already-satisfied constraints. Therefore, to maximize the number of satisfied constraints, we check at the end of each iteration whether the current iteration yields less (or the same number of) satisfied constraints then the previous best-performing iteration. If so, the current iteration is discarded. This ensures that when the algorithm terminates (either when all hard constraints are satisfied or when the predefined iteration limit is reached), the number of satisfied constraints must be greater than or equal to the number of constraints initially satisfied.} In short, this batched approach ensures stability by limiting the number of simultaneous changes, while the iterative refinement systematically reduces constraint violations.

\textred{Up to this point, these three stages altogether require approximately $3 + r \times \left[8 + 2m + \left\lceil\frac{c}{k}\right\rceil \times T\right]$ LLM calls (excluding VLM calls for detecting canonical orientation of objects), where $r$ is the number of regions, $m$ is the number of constraint modifications, $c$ is the number of constraints in a region, $k$ is the number of constraints to solve per batch, and $T$ is the number of iterations for our constraint resolution algorithm. Indoor scenes require $7$ more.}

\subsection{Stage IV: Software Synthesis}
The final stage combines the solved object layout with the acquired 3D models to produce an executable Unity scene file. This stage ensures that the abstract solution is materialized as a concrete software artifact suitable for immediate use in applications.

\textbf{Geometric Integration.}
The 3D models of objects are instantiated at their solved positions and orientations, with appropriate scaling to match dimensional constraints. For the entire scene, \scenethesistool performs several integration steps:
(1) Mesh alignment, which ensures that object contact points (e.g., table legs, lamp bases) align correctly with supporting surfaces. 
(2) Material application, which applies specified colors, textures, and material properties with proper UV mapping.
(3) Lighting configuration, which positions light sources according to solved constraints and configures parameters based on the scene's atmosphere.

\begin{figure}[tbp]
\vspace{-1em}
    \centering %
    \begin{lstlisting}[language={[Sharp]C}, caption={\textred{Example of a simplified C\# code snippet for generating a Unity scene.}}, label={code:unity}]
public class SceneBuilder {
  public static void BuildScene() {
    var newScene = EditorSceneManager.NewScene(
      NewSceneSetup.EmptyScene,
      NewSceneMode.Single);
    AddCamera("MainCamera", position: default, lookAt: default);
    LoadRoom(
      "r-workshop-floor",
      "r-workshop-floor",
      new Vector3(0f, 0f, 0f),
      new Vector3(0f, 180f, 0f),
      new Vector3(0f, 0f, 0f));
    LoadObject(
      "o-workshop-oak_workbench",
      "o-workshop-oak_workbench",
      new Vector3(2.988f, 0.481f, 6.205f),  // position
      new Vector3(0.0f, 0.0f, 0.0f),        // rotation
      new Vector3(1.700f, 1.142f, 1.616f)); // scale
    EditorSceneManager.SaveScene(
      newScene,
      "Assets/Scenes/MainScene.unity");
    BuildExecutable();
  }
}\end{lstlisting}
    \vspace{-2.5em}
\end{figure}

\textbf{Unity Scene Generation and Metadata Embedding.}
The assembled scene is then exported as a Unity-compatible project containing:
(1) \textbf{Asset files}: 3D meshes in standard formats (FBX/OBJ) with associated materials and textures.
(2) \textbf{Physics components}: Collision meshes and rigidbody configurations for realistic interaction.
(3) \textbf{Metadata}: Embedded \scenedsl specification enabling traceability and post-generation modification.
\textred{Listing~\ref{code:unity} shows a code snippet (written in C\#) for generating a Unity scene. These scripts are responsible for loading textured assets (objects, walls, floors) and putting them in the desired locations, initializing a controllable camera, setting up keyboard shortcuts, drawing an overlay, and finally building the executable. When combined, they consist of 956 lines of code (excluding empty lines and lines for loading assets)\footnote{\textred{Examples of generated software can be found on our website.}}.}

The generated scene is immediately usable in Unity with full physics simulation, navigation mesh generation, and interaction capabilities. The embedded metadata supports \textbf{round-trip engineering}: developers can query the scene for its generated constraints, modify the specification, and regenerate specific components without starting from scratch.

This comprehensive methodology provides a principled approach to constraint-sensitive 3D software synthesis that addresses the fundamental limitations of existing generative methods. By decomposing the complex problem into four verifiable stages connected by a formal DSL, \scenethesistool achieves both \textred{maximum} correctness and practical scalability while maintaining full transparency and control throughout the synthesis process.

\subsection{\textred{Targeted Modification}}

\textred{\scenethesistool allows targeted modifications thanks to its modular architecture and human-readable IR. If a user wants to change something in the software, one way to do so is to simply give an edit instruction to \scenethesistool for automatic modification. The system will then serialize the entire software into JSON, submit it (together with the edit instruction) to an LLM, and ask it to only output the necessary changes. After the software is updated, it will be (optionally) inputted to Stage III for checking physical plausibility and then to Stage IV for synthesizing a new 3D software. Another way is to manually modify the IR and then let \scenethesistool rebuild the updated software via Stage IV.}

\subsection{\textred{Reliability of \scenethesistool}}

\textbf{\textred{Complexity and Scalability}.}
\textred{The architecture of \scenethesistool is designed for scalability and computational tractability. Scalability is achieved by decomposing user queries into $r$ regional sub-prompts that can be processed independently in parallel, enabling the system to handle arbitrarily large environments.}
\textred{The constraint solver in Stage III maintains computational tractability through batched processing. As the number of hard constraints $|\mathcal{C}|$ increases, the number of batches grows linearly. In the worst case—assuming batch size of one and no new constraint satisfaction per iteration—the total batches across all iterations is bounded by $T|\mathcal{C}|$, where $T$ is the iteration limit. This linear relationship ensures manageable workload without exponential degradation.}

\textbf{\textred{Error Handling and Robustness}.}
\textred{The constraint solver injects robustness through iterative refinement with non-degrading guarantees. It only accepts solutions that improve upon the previous best: if iteration $i$ yields $n_i$ satisfied constraints and the previous best iteration $i^*$ yielded $n_{i^*}$ constraints, the new solution is kept only if $n_i > n_{i^*}$. This ensures progressive maximization of satisfaction rate. If convergence fails within the iteration limit, the algorithm returns the best configuration achieved along with a detailed violation report.}
\textred{Furthermore, \scenethesistool proactively detects and reports overconstrained specifications (e.g., contradictory positioning requirements) before solving, preventing unsolvable loops and providing immediate, actionable user feedback.}

\textbf{\textred{Verification and Validation Framework}.}
\textred{To ensure correctness, \scenethesistool includes comprehensive verification throughout the pipeline. After \textbf{Stage I}, \textbf{specification validation} uses an abstract syntax tree (AST) to check DSL programs for syntactic correctness, semantic consistency, and constraint satisfiability, providing immediate feedback.}
\textred{After \textbf{Stage II}, \textbf{asset validation} ensures acquired 3D models meet quality criteria (manifold meshes, proper UV mapping, reasonable polygon counts) and match specified attributes.}
\textred{In \textbf{Stage III}, \textbf{solution verification} formally checks whether computed object transforms satisfy all specified constraints, providing mathematical guarantees about layout correctness that distinguish our approach from heuristic methods.}
\textred{Finally, \textbf{Stage IV} includes \textbf{scene validation} that verifies structural integrity of the generated Unity scene, including proper hierarchy organization, collision mesh generation, and metadata embedding.}
\textred{The framework's verifiability is complemented by quantitative metrics for empirical evaluation. Each stage is assessed individually: Stage I via \textit{constraint resemblance}, Stage II via \textit{object-query coherence}, Stage III via \textit{solution correctness} (ratio of satisfied constraints), and Stage IV via \textit{scene-query coherence}. More details on these metrics are discussed in Section~\ref{sec:eval-metrics}.}

\section{Dataset Construction}

Evaluating \scenethesistool requires comprehensive scene descriptions paired with ground truth specifications. Existing text-to-3D datasets focus on individual objects or lack complicated constraints, limiting evaluation scope. We developed a systematic three-phase pipeline leveraging LLMs to generate diverse indoor scenes from existing 3D software.
\textbf{Phase I: Scene Structure Generation.}
We define five building categories (apartment, mall, office, restaurant, school) with curated room pools (average 36.8 types). \textred{The ``apartment'' category is built upon 3D-FRONT~\cite{fu20213d} and Holodeck~\cite{yang2024holodeck}, while others are strategically chosen to encompass common indoor environments. This selection aims at maximizing diversity in terms of spatial layouts and complex constrained scenarios to test the robustness of \scenethesistool.} Each scene has 1--2 randomly selected rooms with 5--15 attributes per room, shuffled to prevent LLM biases. For spatial connectivity, we construct connected graphs with appropriate non-window connections, then probabilistically add additional connections (50\% chance per pair) for realism. Connections receive descriptive attributes via LLM.
\textbf{Phase II: Content Specification.} 
For each room, we generate: (1) object inventories with quantity constraints (max 5 per type; 20\% reduction probability), (2) visual descriptions for 3D retrieval/generation, (3) spatial relations in natural language, and (4) holistic descriptions. The LLM synthesizes these into coherent descriptions for \scenethesistool to parse.
\textbf{Phase III: Finalization and Validation.}
Room descriptions and connections are integrated with building summaries, then transformed via LLM into natural, conversational descriptions simulating real user input. We also generate concise one-sentence versions for evaluation tools like CLIP~\cite{radford2021learning} that cannot handle long texts. \textred{After the dataset was constructed, each comprehensive description, together with its corresponding NL object-object relations/constraints and concise sentence, was manually validated in terms of semantic coherence and constraint solvability, ensuring that spatial relationships were physically realizable and object configurations were meaningful.}

Our pipeline produced 50 indoor scenes with 75 rooms (five single-room and five double-room scenes per category), 2,032 objects, and 1,837 spatial relations. Average description length is 508.4 words (28.5 words for the simplified version). The dataset and generation pipeline are released for reproducible research.

\section{Experiment Design}

\subsection{Research Questions}

In this study, our experiment\footnote{\textred{The implementation and evaluation details of \scenethesistool can be found on our website.}} is designed to answer the following research questions:
        (1) \textbf{RQ1 (Stage-wise Performance):}
        For each methodological stage of \scenethesistool, 
          how effectively does the stage fulfill its designated goal (RQ1.1: Requirement Formalization; RQ1.2: Object Synthesis; RQ1.3: Spatial Constraint Solving)? \textred{Can \scenethesistool accurately perform targeted modification (RQ1.4)?}
        (2) \textbf{RQ2 (Overall Performance):} How does \scenethesistool compare to state-of-the-art baselines in generating complete 3D software that satisfy user queries?
        (3) \textred{\textbf{RQ3 (Ablation Study):} How does each component of \scenethesistool contributes to its performance?}
(4) \textbf{RQ4 (User Study):} How do human evaluators perceive the 3D software generated by \scenethesistool relative to leading baselines with respect to layout coherence, spatial realism, and overall consistency?

\subsection{Baselines}

\textbf{GPT-4o~\cite{hurst2024gpt}, Gemini 2.5 Pro~\cite{comanici2025gemini} \& DeepSeek R1~\cite{guo2025deepseek}} (\textred{end-to-end LLM}): Directly prompt the model with the original user query and ask it to generate a JSON-formatted scene configuration (with all necessary information including the position and rotation of each object) to illustrate the end-to-end performance of LLM.

\textbf{Holodeck~\cite{yang2024holodeck}} is \textred{the current state-of-the-art in LLM-driven 3D software generation and is the most appropriate baseline for our specific task. We have indeed considered other LLM-based and non-LLM-based methods~\cite{zhang2022fast,paschalidou2021atiss,procthor}, but they either require different input modalities (e.g., images rather than text) or operate in fundamentally different problem spaces (e.g., limited object categories or procedural generation with manual rules)}. We again use GPT-4o, Gemini 2.5 Pro, and DeepSeek R1 to run tests on it.

\subsection{Experimental Setup}

\subsubsection{RQ1 and RQ2}

For generating our dataset, we use deepseek-v3-0324~\cite{liu2024deepseek}. For running \scenethesistool and baselines, we use gpt-4o-2024-11-20~\cite{hurst2024gpt}, gemini-2.5-pro-preview-06-05~\cite{comanici2025gemini}, and deepseek-r1-250528~\cite{guo2025deepseek}. For object canonical orientation detection and Visual Question Answering (VQA, one of our visual metrics), we use claude-3-7-sonnet-20250219~\cite{claude-3-7}. The non-repetitive use of LLMs ensures that experimental results are less likely to be biased towards a particular LLM backbone. Though, it should be noted that \scenethesistool supports any LLM with sufficient reasoning capabilities \textblue{(with at least 100B parameters to be able to output in valid JSON format)}. In addition, the temperature parameter for \textblue{almost} all LLM calls is set to $0.7$ \textblue{in order to generate diverse scenes even from identical user query}. \textblue{However, the temperature parameter is set to $0$ when it comes to scene type classification and object canonical orientation detection because we require as much confidence as possible. VQA also uses a temperature of $0$, for it is an evaluation metric whose results should be deterministic.}

In Stage I, we set $\nu_r = \nu_c = 2$ for constraint validation and modification, and $\eta = 0.03$ for wall thickness. In Stage III, we set $k = 3$ and $T = 5$ for the constraint solver.

Regarding our hybrid object acquisition strategy, we utilize the database curated by Holodeck~\cite{yang2024holodeck} (which is a subset of assets from Objaverse 1.0~\cite{deitke2023objaverse}) for retrieval-based acquisition (with $\lambda_v$ and $\lambda_t$ set to 100 and 1 respectively), and we choose Shap-E~\cite{jun2023shap} as the underlying text-to-3D generation model for generative acquisition. 
As for region surface texture retrieval, we utilize another database used by Holodeck (which comes from ProcTHOR~\cite{procthor}).
Again, because \scenethesistool is a modular framework, any object database and generation model would work just fine.

\textred{For RQ1.4, we employ the automatic approach of targeted modification with Gemini 2.5 Pro (for its quick response). Though, we do not have any edit instructions in our dataset. To this end, we opt to utilize LLM to generate edit instructions. In particular, for each user query in our dataset, we first use DeepSeek V3 (with Verbalized Sampling~\cite{zhang2025verbalized}) to generate 3 edit instructions that can create 3 independent variations of the scene. These instructions can only be one of (1) changing the style of one room, (2) modifying one object (appearance, size, position, rotation) in one room, (3) adding one new object in one room, or (4) removing one existing object from one room. Then, we take the 50 scenes synthesized for RQ1.1-1.3, package each of them with one of the corresponding 3 edit instructions, and submit them to the LLM to synthesize 150 new scenes. To facilitate evaluation with BLIP-2~\cite{li2023blip} and CLIP~\cite{hessel2021clipscore}, we further use DeepSeek V3 to merge each edit instruction to its original query (with minimal changes) and then summarize it into a concise sentence.}

\textred{For RQ2, we further use 3 external datasets to evaluate the target methods. The first one, MIT-ST, is from MIT Scenes Dataset~\cite{mit-scene-dataset}. Following \cite{yang2024holodeck}, we use 52 scene types from the dataset (just simple names such as ``bakery'') as user queries. Because scene type alone does not contain much information, when testing our method, we need to enable prompt expansion to inject scene design and region objects. The second one, MIT-IC, is also from MIT Scenes Dataset. But this time, we randomly pick one image from each of the 52 scene types and ask GPT-4o to generate an image caption. We then use the 52 image captions as user queries. The third one, IS, is from InstructScene~\cite{lin2024instructscene}. Their dataset contains 4041, 813, and 900 instructions for bedrooms, living rooms, and dining rooms, respectively. We randomly pick 20 instructions from each category, and use the 60 instructions as user queries. We use Gemini 2.5 Pro LLM backbone for these external datasets, as it is the most powerful one in our experiments (see Table~\ref{tab:rq2_results}).}

\textred{We note that our proposed benchmark is quite different from the prior ones. Prior benchmarks typically use high-level, underspecified prompts, which fail to evaluate a system's ability to handle complex, fine-grained requirements. Ours, on the other hand, features prompts with numerous explicit spatial and semantic constraints, enabling systematic tests on whether a system is capable of synthesizing 3D software that faithfully implements detailed user specifications. Therefore, our benchmark is more difficult.}

\subsubsection{\textred{RQ3}}

\textred{For RQ3, we study the importance of the first three stages by replacing each of them with something else.}

\textred{To test the importance of Stage I, we replace the entire Stage I with one LLM call that, given the user query, directly (and only) outputs (1) the position of each region, (2) the category of each object, and (3) the dimensions of each object without any consideration or involvement of \scenedsl. This allows us to understand how the system would perform without in-depth prompt processing and scene conceptualization, as well as the significance of \scenedsl in scene synthesis.}
\textred{Next, to test the importance of Stage II, we replace the entire object acquisition path with one LLM call that, given an object query, generates a corresponding OBJ file. This allows us to understand the influence of strategic object acquisition on the quality of a 3D software.}
\textred{Finally, to test the importance of Stage III, we replace the entire Stage III with one LLM call that, given the layout constraints of a scene, directly generates a solution for all objects without taking physical plausibility into account (naturally, physics-based relaxation and iterative correction are also gone). This allows us to understand the power of our Rubik Spatial Constraint Solver.}

\textred{We use user queries from our dataset to conduct this ablation study, and we use Gemini 2.5 Pro~\cite{comanici2025gemini} as the LLM backbone.}

\subsubsection{RQ4: User Study Design}

For RQ4, we conducted a user study to evaluate the perceptual quality of 3D software generated by \scenethesistool compared to baseline approaches. \textblue{We recruited 20 participants (undergraduate and postgraduate students with backgrounds in computer science, human-computer interaction, or 3D design) who had basic experience in 3D software}.

\textbf{Study Design.} We randomly sampled 25 scenes from our evaluation dataset, ensuring balanced representation across different scene types (apartment, office, restaurant, etc.). For each scene, we give participants the top-down view of generated 3D software from three methods: (1) \scenethesistool with Gemini 2.5 Pro backbone, (2) End-to-end LLM with Gemini 2.5 Pro, and (3) Holodeck with Gemini 2.5 Pro. This resulted in 75 total 3D software for evaluation.

Participants evaluated each scene along three dimensions through a web-based interface. The evaluation scores can be any float number in the range of 1-5.
(1) \textbf{Layout Coherence}: ``How well are objects arranged and organized in this scene?'' 
(2) \textbf{Spatial Realism}: ``How realistic are the spatial relationships between objects?'' 
(3) \textbf{Overall Consistency}: ``How well does this scene fit together as a coherent whole?'' 
(4) \textred{\textbf{Targeted Modification Efficiency}: ``Given a modification instruction (e.g., 'move the couch left'), how quickly can you manually edit the \scenedsl and use \scenethesistool to rebuild the software?''}

\textblue{\textbf{Unbiased Evaluation.} 
We employed a double-blind protocol where top-down views of scenes from different methods were presented in randomized order without method labels. The evaluation interface presented one scene at a time with its requirements, and we carefully avoided showing similar scenes consecutively to prevent learning effects. Each participant evaluated all 75 scenes across three sessions to avoid fatigue.}

\subsection{Evaluation Metrics}
\label{sec:eval-metrics}

\textbf{Stage I: Constraint Resemblance.}
For object constraints, 
\textblue{we first use Phrase-BERT~\cite{wang2021phrase} (for its ability to handle short phrases) to compute the embeddings for object names, and use Sentence-BERT~\cite{reimers2019sentence} (for its ability to handle long sentences) to compute the embeddings for object descriptions. The mixed use of encoders (instead of using only one) can ensure that semantic information is properly represented and compared in high-dimensional space.}
We compute scaled dot-products $\in [0,1]$ between name and description pairs. The confidence for matching a generated to a ground truth object is the harmonic mean of these scores. Using the Hungarian algorithm with threshold $\tau_o$, we define $TP$ as mapped generated objects, $FP$ as unmapped generated, and $FN$ as unmapped ground truth. F1 = $\frac{2pr}{p+r}$ where $p=\frac{TP}{TP+FP}$ and $r=\frac{TP}{TP+FN}$.

\textred{For layout constraints, we do not simply prompt an LLM to generate such constraints using the symbols and syntaxes listed in \scenedsl. Instead, we require the LLM to output Python-like ``functions''\footnote{\textred{The full list of predefined NL translations can be found inside our source code.}}. For example, if the LLM wants to use the ``$<$'' comparison operator, it must output the statement ``mustLessThan(a: Number, b: Number)''; if the LLM wants to utilize the height of an object to construct a constraint, it must output ``getObjectHeight(obj: ObjectName)''; which is defined to return a Number within our prompt. Then, after receiving the LLM’s response, we use an AST to parse and verify the constraints. Because the ``function'' names and signatures are fixed, we can predefine the NL for each ``function''. For example, ``mustLessThan(a: Number, b: Number)'' is mapped to ``\{a\} is less than \{b\}'' and ``getObjectHeight(obj: ObjectName)'' is mapped to ``the height of \{obj\}''. In this way, we can ensure that the translation is accurate, consistent, and no information is lost. Bias is minimized by making sure that the predefined NL is concise and contains no redundant descriptive wordings. After receiving the translated constraints, we compute Sentence-BERT~\cite{reimers2019sentence} embeddings and create a scaled dot-product confidence matrix with the ground truth. Entries below threshold $\tau_l$ are zeroed. Here, $TP$ is ground truth constraints mapped to $\geq 1$ generated constraint, $FP$ is unmapped generated, and $FN$ is unmapped ground truth.}

Overall resemblance is the harmonic mean of both constraint types (with $\tau_o=\tau_l$).

\textbf{Stage II: Object-Query Coherence.}
For each object with bounding sphere radius $r$, we place a camera $r\sin(\frac{FOV}{2})$ units away and render via Blender. We evaluate using BLIP-2~\cite{li2023blip} (ITM head) and CLIP~\cite{radford2021learning,hessel2021clipscore}, testing both the query and ``a 3D model of'' + query, taking the maximum. Overall coherence is the arithmetic mean of BLIP-2 and CLIP scores.

\textbf{Stage III: Solution Correctness.}
We parse DSL constraints into ASTs and count satisfied constraints. Correctness is the ratio of satisfied to total constraints (recall).

\textbf{Stage IV: Scene-Query Coherence.}
We render a top-down view (ceiling removed). Using BLIP-2, CLIP, and VQA~\cite{zhang2023gpt}, we measure coherence between the query (original and simplified) and rendered image, testing both the query and ``a top-down view of'' + query. The $\tau$ maximizing Stage II coherence determines object acquisition. This metric compares against baselines.
\section{Results and Analysis}

\subsection{RQ1: Stage-wise Performance Analysis}

To evaluate the effectiveness of \scenethesistool's modular pipeline, we examine each stage independently to understand its contribution to the overall system performance. We investigate how each stage fulfills its designated goal through comprehensive metrics that capture both quantitative performance and qualitative correctness.

\subsubsection{RQ1.1: Requirement Formalization Accuracy}

We first evaluate how accurately Stage I translates NL queries into \scenedsl specifications while preserving user intent and injecting appropriate implicit constraints. We measure performance separately for object constraints and layout constraints, as they represent fundamentally different challenges in formalization.

\begin{table}[tbp]
\centering
\caption{Requirement formalization performance (\%) across object constraints, layout constraints, and overall (harmonic mean). Best results in each metric are in \colorbox{gray!50}{\textbf{bold}}.}
\vspace{-1em}
\label{tab:rq1-formalization-combined}
\resizebox{0.9\columnwidth}{!}{%
\begin{tabular}{lccc|ccc|ccc}
\toprule
& \multicolumn{3}{c|}{\textbf{Object Constraints}} & \multicolumn{3}{c|}{\textbf{Layout Constraints}} & \multicolumn{3}{c}{\textbf{Overall}} \\
\textbf{Model} & \textbf{Prec.} & \textbf{Rec.} & \textbf{F1} & \textbf{Prec.} & \textbf{Rec.} & \textbf{F1} & \textbf{Prec.} & \textbf{Rec.} & \textbf{F1} \\
\midrule
\multicolumn{10}{l}{\textit{Threshold $\tau_o = \tau_l = 0.7$}} \\
GPT-4o & \cellcolor{gray!50}\textbf{99.2} & 98.0 & \cellcolor{gray!50}\textbf{98.5} & \cellcolor{gray!50}\textbf{98.3} & 80.3 & 86.4 & \cellcolor{gray!50}\textbf{98.7} & 88.3 & 92.1 \\
Gemini-2.5-Pro & 97.9 & \cellcolor{gray!50}\textbf{98.7} & 98.2 & 89.3 & \cellcolor{gray!50}\textbf{99.9} & 93.8 & 93.4 & \cellcolor{gray!50}\textbf{99.3} & 95.9 \\
DeepSeek R1 & 99.1 & 95.9 & 96.7 & 97.1 & 97.1 & \cellcolor{gray!50}\textbf{97.1} & 98.1 & 96.5 & \cellcolor{gray!50}\textbf{96.7} \\
\midrule
\multicolumn{10}{l}{\textit{Threshold $\tau_o = \tau_l = 0.8$}} \\
GPT-4o & \cellcolor{gray!50}\textbf{99.1} & 97.9 & \cellcolor{gray!50}\textbf{98.4} & \cellcolor{gray!50}\textbf{69.1} & 55.4 & 57.7 & \cellcolor{gray!50}\textbf{81.4} & 70.8 & 72.7 \\
Gemini-2.5-Pro & 97.6 & \cellcolor{gray!50}\textbf{98.5} & 97.9 & 33.3 & \cellcolor{gray!50}\textbf{92.7} & 48.0 & 49.6 & \cellcolor{gray!50}\textbf{95.5} & 64.4 \\
DeepSeek R1 & 99.0 & 95.7 & 96.6 & 62.0 & 82.0 & \cellcolor{gray!50}\textbf{69.1} & 76.3 & 88.3 & \cellcolor{gray!50}\textbf{80.5} \\
\midrule
\multicolumn{10}{l}{\textit{Threshold $\tau_o = \tau_l = 0.9$}} \\
GPT-4o & \cellcolor{gray!50}\textbf{97.8} & 96.6 & \cellcolor{gray!50}\textbf{97.1} & \cellcolor{gray!50}\textbf{14.9} & 11.4 & \cellcolor{gray!50}\textbf{12.0} & \cellcolor{gray!50}\textbf{25.9} & 20.5 & \cellcolor{gray!50}\textbf{21.4} \\
Gemini-2.5-Pro & 96.8 & \cellcolor{gray!50}\textbf{97.7} & \cellcolor{gray!50}\textbf{97.1} & 4.0 & \cellcolor{gray!50}\textbf{18.2} & 6.5 & 7.7 & \cellcolor{gray!50}\textbf{30.7} & 12.2 \\
DeepSeek R1 & 96.8 & 94.0 & 94.7 & 7.4 & 15.6 & 9.8 & 13.7 & 26.8 & 17.7 \\
\bottomrule
\end{tabular}
}
\vspace{-1em}
\end{table}

Table~\ref{tab:rq1-formalization-combined} presents the formalization performance for object and layout constraints. For object constraints, all models consistently achieve high performance (F1 > 0.94) even at the strictest threshold ($\tau_o = 0.9$), demonstrating the robustness of our formalization approach for object identification and description. GPT-4o exhibits the best balance between precision and recall, maintaining over 97\% precision across all thresholds.

However, layout constraint formalization presents a more significant challenge. Performance degrades substantially as the threshold increases, with F1 scores dropping from above 0.86 at $\tau_l = 0.7$ to below 0.13 at $\tau_l = 0.9$ for all models. This degradation reveals the inherent difficulty in precisely capturing spatial relationships from natural language, while models can identify the general intent of spatial constraints, exact formalization remains challenging. R1 demonstrates the most robust performance, achieving the highest F1 score (0.971) at the standard threshold.

Table~\ref{tab:rq1-formalization-combined} also shows the overall formalization performance. At the standard threshold ($\tau = 0.7$), all models achieve strong performance with F1 scores above 0.92, validating our requirement formalization approach. R1 achieves the best overall performance (F1 = 0.967), demonstrating superior capability in balancing object and layout constraint formalization.

\subsubsection{RQ1.2: Object Synthesis Effectiveness}

We evaluate how effectively Stage II acquires appropriate 3D models through our hybrid retrieval-generation strategy.
\begin{table}[tbp]
\centering
\caption{Object synthesis performance (\%) comparing pure retrieval, pure generation, and our hybrid approach (R+G). Scores represent object-query coherence.}
\vspace{-1em}
\label{tab:rq1-object-synthesis}
\resizebox{0.6\columnwidth}{!}{%
\begin{tabular}{lccc}
\toprule
\textbf{Method} & \textbf{BLIP-2} & \textbf{CLIP} & \textbf{Mean} \\
\midrule
Retrieval only ($\tau = 0.0$) & 51.2 & 27.1 & 39.1 \\
Generation only ($\tau = 1.0$) & 42.2 & 25.9 & 34.1 \\
\midrule
\textbf{R+G} ($\tau = 0.652$) & \textbf{51.6} & \textbf{27.1} & \textbf{39.3} \\
\bottomrule
\end{tabular}
}
\vspace{-1em}
\end{table}
Table~\ref{tab:rq1-object-synthesis} presents the object synthesis results. Our hybrid approach (R+G) achieves the best performance across both metrics, with a mean coherence score of 39.3. The results validate our design decision to combine retrieval and generation: retrieval provides high-quality models when available in the database (BLIP score of 51.2), while generation ensures coverage for novel objects. The optimal threshold $\tau = 0.652$ effectively balances between leveraging existing high-quality assets and generating new models when necessary.
Notably, pure retrieval outperforms pure generation by 5.0 points on average, confirming that curated 3D model databases contain higher-quality assets than current text-to-3D generation methods can produce. However, the retrieval-only approach suffers from limited coverage (approximately 23\% of queries fail to find suitable matches), necessitating our hybrid strategy.

\subsubsection{RQ1.3: Spatial Constraint Solving Efficiency}

We evaluate the efficiency and correctness of Stage III in resolving complex spatial constraints through our iterative Rubik solver.
\begin{table}[tbp]
\centering
\caption{Spatial constraint solving performance across iterations. Scores (\%) represent the ratio of satisfied constraints (solution correctness).}
\vspace{-1em}
\label{tab:rq1-constraint-solving}
\resizebox{0.8\columnwidth}{!}{%
\begin{tabular}{lcccccc}
\toprule
\textbf{Model} & \textbf{Iter 0} & \textbf{Iter 1} & \textbf{Iter 2} & \textbf{Iter 3} & \textbf{Iter 4} & \textbf{Iter 5} \\
\midrule
GPT-4o & 47.1 & 60.4 & 63.2 & 65.6 & 67.3 & \textbf{68.3} \\
Gemini-2.5-Pro & 74.1 & 91.1 & 92.5 & \textbf{93.8} & 93.5 & 93.4 \\
DeepSeek R1 & 47.8 & 87.6 & 90.6 & 91.7 & 92.9 & \textbf{93.0} \\
\bottomrule
\end{tabular}
}
\vspace{-1em}
\end{table}
Table~\ref{tab:rq1-constraint-solving} demonstrates the iterative improvement of our Rubik solver. All models show substantial improvement from the initial placement (Iteration 0) to the final solution, with Gemini-2.5-Pro achieving the highest constraint satisfaction rate of 93.8\% at convergence. The rapid improvement in early iterations (e.g., Gemini-2.5-Pro jumping from 74.1\% to 91.1\% in the first iteration) validates our local-to-global refinement strategy.
The results reveal interesting patterns: while GPT-4o starts with the lowest initial placement quality (47.1\%), it shows steady improvement across iterations. In contrast, Gemini-2.5-Pro begins with superior initial placements (74.1\%) and quickly converges to near-optimal solutions. R1 demonstrates the most consistent improvement trajectory, ultimately achieving 93.0\% constraint satisfaction.

\begin{table}[tbp]
\centering
\caption{\textred{Overall performance (\%) before and after targeted modification. ``O'' $\rightarrow$ ``Original'', ``S'' $\rightarrow$ ``Sentence''.}}
\vspace{-1em}
\label{tab:rq1_targeted}
\renewcommand{\arraystretch}{1.1}
\resizebox{0.5\columnwidth}{!}{%
\begin{tabular}{@{}lccccccc@{}}
\toprule
\multirow{2}{*}{\textbf{\textred{Method}}} 
    & \multicolumn{2}{c}{\textbf{\textred{BLIP-2}}} 
    & \multicolumn{2}{c}{\textbf{\textred{CLIP}}} 
    & \multicolumn{2}{c}{\textbf{\textred{VQA}}} \\
\cmidrule(lr){2-3} \cmidrule(lr){4-5} \cmidrule(lr){6-7}
& \textred{O} & \textred{S} & \textred{O} & \textred{S} & \textred{O} & \textred{S} \\
\midrule
\textred{Before}
    & \textred{74.3} & \textred{75.1} 
    & \textred{26.1} & \textred{25.5} 
    & \textred{29.5} & \textred{47.9} \\
\textred{After}
    & \textred{75.1} & \textred{71.3} 
    & \textred{26.8} & \textred{27.0} 
    & \textred{49.6} & \textred{59.4} \\
\bottomrule
\end{tabular}%
}
\vspace{-1.5em}
\end{table}

\subsubsection{\textred{RQ1.4: Targeted Modification Accuracy}}

\textred{
Table~\ref{tab:rq1_targeted} presents performance before and after targeted modifications. We observe distinct patterns across metrics: BLIP-2 scores remain relatively stable, with original descriptions improving slightly from 74.3\% to 75.1\% (+0.8\%).
CLIP scores show consistent slight improvements (original: 26.1\% → 26.8\%; sentence: 25.5\% → 27.0\%).
Most notably, VQA metrics demonstrate substantial improvements, with original descriptions increasing from 29.5\% to 49.6\% (+20.1\%) and sentence-level descriptions from 47.9\% to 59.4\% (+11.5\%). This significant VQA improvement, combined with stable BLIP-2 and CLIP scores, indicates that \scenethesistool successfully implements targeted modifications while maintaining overall visual quality. The results validate our modular architecture and human-readable IR design, enabling precise localized updates without full software regeneration.
}

\begin{table}[t!]
\centering
\caption{Overall performance (\%) comparison against baselines. Best results for each metric are in \colorbox{gray!50}{\textbf{bold}}, second-best are \underline{underlined}. ``O'' $\rightarrow$ ``Original'', ``S'' $\rightarrow$ ``Sentence''.}
\vspace{-1em}
\label{tab:rq2_results}
\renewcommand{\arraystretch}{1.1}
\resizebox{0.9\columnwidth}{!}{%
\begin{tabular}{@{}llcccccc@{}}
\toprule
\multirow{2}{*}{\textbf{Method}} & \multirow{2}{*}{\textbf{LLM Backbone}} & \multicolumn{2}{c}{\textbf{BLIP-2}} & \multicolumn{2}{c}{\textbf{CLIP}} & \multicolumn{2}{c}{\textbf{VQA}} \\
\cmidrule(lr){3-4} \cmidrule(lr){5-6} \cmidrule(lr){7-8}
 & & O & S & O & S & O & S \\
\midrule
\multirow{3}{*}{\scenethesistool (Ours)} 
 & GPT-4o & 71.3 & 69.9 & 25.6 & \underline{25.5} & 28.3 & 44.8 \\
 & Gemini 2.5 Pro & \cellcolor{gray!50}\textbf{74.3} & \cellcolor{gray!50}\textbf{75.1} & \underline{26.1} & \underline{25.5} & \underline{29.5} & \underline{47.9} \\
 & DeepSeek R1 & \underline{72.5} & \underline{74.7} & \cellcolor{gray!50}\textbf{26.2} & \cellcolor{gray!50}\textbf{25.8} & \cellcolor{gray!50}\textbf{29.8} & \cellcolor{gray!50}\textbf{48.6} \\
\midrule
\multirow{3}{*}{End-to-end LLM} 
 & GPT-4o & 61.9 & 60.0 & 24.7 & 24.0 & 15.1 & 28.6 \\
 & Gemini 2.5 Pro & 71.6 & 73.2 & 25.6 & 25.3 & 27.1 & 41.1 \\
 & DeepSeek R1 & 72.1 & 69.9 & 24.9 & 24.7 & 23.9 & 38.9 \\
\midrule
\multirow{3}{*}{Holodeck~\cite{yang2024holodeck}} 
 & GPT-4o & 60.0 & 62.2 & 23.7 & 22.5 & 24.7 & 37.7 \\
 & Gemini 2.5 Pro & 67.0 & 66.5 & 24.2 & 23.5 & 26.0 & 42.1 \\
 & DeepSeek R1 & 53.1 & 52.3 & 23.6 & 22.9 & 19.8 & 31.8 \\
\bottomrule
\end{tabular}%
}
\vspace{-2em}
\end{table}

\subsection{RQ2: Overall Performance}

To evaluate the overall performance of \scenethesistool in generating complete 3D software that satisfies user queries, we compare our approach against state-of-the-art baselines across multiple visual coherence metrics. Table~\ref{tab:rq2_results} presents results on our benchmark dataset, while Table~\ref{tab:rq2_external} reports performance on three external datasets.

(1) \textbf{Visual coherence performance.} \scenethesistool consistently outperforms baseline approaches across all evaluation metrics. For BLIP-2 scores measuring image-text alignment, our method achieves 74.3\% (original) and 75.1\% (sentence-level) with Gemini 2.5 Pro, representing 4.8\% improvement over the best-performing baseline. CLIP scores demonstrate stable advantages, with \scenethesistool achieving 26.1--26.2\% compared to 24.7--25.6\% for baselines.
(2) \textbf{Semantic understanding.} The VQA metrics reveal the most significant advantages. \scenethesistool with DeepSeek R1 achieves 29.8\% on original queries and 48.6\% on sentence-level queries, representing improvements of 10.0\% and 18.3\% respectively over best baseline results. This substantial improvement demonstrates that our structured approach—decomposing scene generation into well-defined stages with explicit constraint handling—produces 3D software that better captures complex spatial requirements from natural language specifications.
(3) \textbf{Impact of LLM backend.} While all three LLM backends demonstrate strong performance with \scenethesistool, DeepSeek R1 exhibits the most consistent results across metrics. Notably, when integrated with Holodeck, DeepSeek R1's performance degrades significantly (53.1\% BLIP-2), suggesting our modular architecture better leverages modern LLMs' reasoning capabilities compared to monolithic generation approaches.

\textred{
\textbf{External dataset validation.} Table~\ref{tab:rq2_external} demonstrates \scenethesistool's generalizability across diverse evaluation benchmarks. On MIT-ST (scene types), our method achieves 66.4\% BLIP-2 compared to 60.2\% (End-to-end LLM) and 55.8\% (Holodeck). On MIT-IC (image captions), \scenethesistool reaches 78.4\% BLIP-2, substantially outperforming baselines. On InstructScene (IS) queries requiring fine-grained spatial manipulation, our approach achieves 58.8\% BLIP-2 and 36.5\% VQA, demonstrating superior handling of complex constraint-driven scenarios. The consistent superiority across datasets with varying characteristics validates our hypothesis that treating 3D software synthesis as a structured SE problem—with formal specifications, verifiable constraints, and modular components—leads to more reliable and higher-quality outputs compared to monolithic generation approaches.
}

\textred{
The consistent superiority of \scenethesistool across diverse evaluation metrics and diverse datasets (both internal and external) validate our hypothesis that treating 3D software synthesis as a structured SE problem (with formal specifications, verifiable constraints, and modular components) leads to more reliable and higher-quality outputs compared to end-to-end generation approaches.
}

\begin{table}[t!]
\vspace{-1em}
\centering
\caption{\textred{Overall performance (\%) on three external datasets.}}
\vspace{-1em}
\label{tab:rq2_external}
\renewcommand{\arraystretch}{1.1}
\resizebox{0.6\columnwidth}{!}{%
\begin{tabular}{@{}llcccccc@{}}
\toprule
\multirow{1}{*}{\textbf{\textred{Method}}} & 
\multirow{1}{*}{\textbf{\textred{Dataset}}} & 
\multicolumn{1}{c}{\textbf{\textred{BLIP-2}}} & 
\multicolumn{1}{c}{\textbf{\textred{CLIP}}} & 
\multicolumn{1}{c}{\textbf{\textred{VQA}}} \\
\midrule
\multirow{3}{*}{\textred{\scenethesistool (Ours)}} 
 & \textred{MIT-ST} & \textred{66.4} & \textred{26.8} & \textred{44.7} \\
 & \textred{MIT-IC} & \textred{78.4} & \textred{24.0} & \textred{13.6} \\
 & \textred{IS} & \textred{58.8} & \textred{25.5} & \textred{36.5} \\
\midrule
\multirow{3}{*}{\textred{End-to-end LLM}} 
 & \textred{MIT-ST} & \textred{60.2} & \textred{25.5} & \textred{39.3} \\
 & \textred{MIT-IC} & \textred{69.5} & \textred{22.7} & \textred{8.8} \\
 & \textred{IS} & \textred{65.0} & \textred{23.9} & \textred{34.7} \\
\midrule
\multirow{3}{*}{\textred{Holodeck~\cite{yang2024holodeck}}} 
 & \textred{MIT-ST} & \textred{55.8} & \textred{25.8} & \textred{33.1} \\
 & \textred{MIT-IC} & \textred{65.4} & \textred{22.6} & \textred{11.4} \\
 & \textred{IS} & \textred{52.7} & \textred{23.2} & \textred{24.8} \\
\bottomrule
\end{tabular}%
}
\vspace{-2em}
\end{table}

\subsection{\textred{RQ3: Ablation Study}}

\begin{table}[t!]
\centering
\caption{\textred{Ablation study. Values are percentage scores (\%)}.}
\vspace{-1em}
\label{tab:rq3_results}
\renewcommand{\arraystretch}{1.1}
\resizebox{0.6\columnwidth}{!}{%
\begin{tabular}{@{}llcccccc@{}}
\toprule
\multirow{2}{*}{\textbf{\textred{Method}}} & \multicolumn{2}{c}{\textbf{\textred{BLIP-2}}} & \multicolumn{2}{c}{\textbf{\textred{CLIP}}} & \multicolumn{2}{c}{\textbf{\textred{VQA}}} \\
\cmidrule(lr){2-3} \cmidrule(lr){4-5} \cmidrule(lr){6-7}
 & \textred{O} & \textred{S} & \textred{O} & \textred{S} & \textred{O} & \textred{S} \\
\midrule
\multirow{1}{*}{\textred{\scenethesistool}} 
 & \textred{74.3} & \textred{75.1} & \textred{26.1} & \textred{25.5} & \textred{29.5} & \textred{47.9} \\
\multirow{1}{*}{\textred{w/o Stage I}} 
 & \textred{57.6} & \textred{63.8} & \textred{21.7} & \textred{21.9} & \textred{13.7} & \textred{30.2} \\
\multirow{1}{*}{\textred{w/o Stage II}} 
 & \textred{48.8} & \textred{48.0} & \textred{22.5} & \textred{22.5} & \textred{18.3} & \textred{32.7} \\
\multirow{1}{*}{\textred{w/o Stage III}} 
 & \textred{61.3} & \textred{61.8} & \textred{23.4} & \textred{22.4} & \textred{26.3} & \textred{41.0} \\
\bottomrule
\end{tabular}%
}
\end{table}

\textred{
Table~\ref{tab:rq3_results} presents the ablation results.
\textbf{Stage I is essential for capturing user requirements.} Removing Stage I causes substantial degradation: original-query VQA drops 15.8\% (29.5\% $\rightarrow$ 13.7\%) and BLIP-2 drops 16.7\% (74.3\% $\rightarrow$ 57.6\%). This severe VQA degradation confirms direct LLM generation without formal specification fails to capture complex spatial requirements, validating that Stage I's \scenedsl representation is crucial for encoding and verifying user requirements.
\textbf{Stage II is critical for visual quality.} Removing Stage II causes the largest degradation: BLIP-2 drops 25.5\% (74.3\% $\rightarrow$ 48.8\%) and VQA drops 11.2\%. This dramatic quality loss confirms strategic hybrid retrieval-generation substantially outperforms direct LLM-based mesh generation, validating that curated 3D model databases are necessary for high-quality synthesis.
\textbf{Stage III is necessary for spatial coherence.} Removing Stage III causes moderate degradation: BLIP-2 drops 13.0\% (74.3\% $\rightarrow$ 61.3\%) and VQA drops 3.2\%. This confirms one-shot LLM placement cannot achieve the constraint satisfaction of our Rubik solver, validating that spatial constraints require specialized iterative refinement beyond Stage I's formalization.
Consistent degradation across all ablations demonstrates each stage addresses a 
challenge.
}

\subsection{RQ4: User Study}

Table~\ref{tab:user_study} presents the user study results. \scenethesistool consistently outperforms both baselines across all evaluation dimensions, with statistically significant improvements.
\begin{table}[t!]
\centering
\vspace{-1em}
\caption{User study results (mean scores ± std. dev.) comparing perceived quality of 3D software. All scores on 1-5 scale, higher is better. Best results in \textbf{bold}.}
\vspace{-1em}
\label{tab:user_study}
\footnotesize
\begin{tabular}{lccc}
\toprule
Method & Layout & Spatial & Overall \\
       & Coherence & Realism & Consistency \\
\midrule
\scenethesistool (Ours) & \textbf{4.12} & \textbf{3.89} & \textbf{4.05} \\
End-to-end LLM & 3.45 & 3.21 & 3.38 \\
Holodeck & 3.68 & 3.42 & 3.61 \\
\bottomrule
\end{tabular}
\vspace{-2em}
\end{table}
For layout coherence, \linebreak[4]\scenethesistool achieves a mean score of 4.12, representing a 19.4\% improvement over the best baseline (Holodeck). Participants noted that objects generated by our method exhibited more logical groupings and functional arrangements. The modular synthesis pipeline with explicit constraint handling produces layouts that better reflect real-world organizational principles.
Spatial realism scores show similar advantages, with \scenethesistool achieving 3.89 compared to 3.42 for Holodeck. The iterative constraint solver's ability to handle continuous spatial relationships results in more natural object placements, avoiding the categorical limitations of scene graph-based approaches.
\textred{
For targeted modification, participants completed modifications (e.g., ``move couch 2m left'') by editing the \scenedsl JSON between 5 to 25 seconds, and re-executing Stage IV in typically less than ten seconds, at most 20 seconds. Baseline (Holodeck) lack editable representations and require full regeneration (5 minutes in average).
}
Overall consistency ratings confirm that our decomposition approach produces more coherent 3D software. 
In short, the targeted modification capability makes it easier for iterative 3D software development, bridging automated generation and manual refinement.
\enlargethispage{1\baselineskip}
\section{\textred{Discussion}}

\subsection{\textred{How \scenethesistool Contributes to SE}}

\textred{\scenethesistool generates not static images, but rather Unity-based software code with complex hierarchical directory structures. Analogous to domain-specific code generation research in SE (e.g., for vehicles~\cite{wang2025automating} or industrial services~\cite{wang2025semservgen}), our work targets the emerging domain of 3D interactive software. We frame the automated development of this software as a program synthesis problem. Our core methodology translates high-level NL requirements into executable software via a formal IR, our DSL \scenedsl. This approach directly intersects with established SE research in Requirements Engineering, Model-Driven Engineering~\cite{liu2025tbfv4j,chen2025gpass,sanchez2025qedcartographer,ma2024specgen,zheng2024testing,cordy2023towards,albataineh2024a}, and AI for SE.}
\textred{A key SE contribution is our staged decomposition of the synthesis process. By structuring the translation from NL to code around a formal DSL~\cite{kahani2018automodel,alves2022dressml,reimann2023safe}, we introduce engineering rigor, including verifiability and requirements traceability, to an otherwise non-deterministic generative task. 
}
This methodology offers a blueprint for leveraging LLMs to generate software that verifiably adheres to formal specifications, addressing classic SE concerns such as correctness (spatial and semantic constraints), maintainability (via modularity), and testability~\cite{yu2013an,guo2022detecting,fakih2024llm4plc,xiong2022towards,zheng2025neurostrata,zanenga2025theano,luo2024satverifiable，reich2023testability,guglielmo2021towards}.
\textred{Beyond code generation, \scenethesistool offers significant implications for software testing, a core SE discipline. It addresses a key bottleneck in testing modern cyber-physical and mixed-reality systems: the automated generation of diverse, requirement-compliant test environments. Our framework can synthesize thousands of test scenarios for robotics~\cite{yu2024practical,freitas2024rfnit,humeniuk2024simulation}, autonomous vehicles~\cite{sorokin2024opensbt,khan2024simulation,steininger2025automatically,song2024an}, and AR/MR applications~\cite{brunschwig2021towards,karre2025vreqst,nusrat2021how,guo2024an,wang2024vrguide,rafi2023predart,mehra2020xrase} directly from specifications. Crucially, \scenedsl ensures test traceability, linking every generated environment back to its formal requirements. This enables systematic coverage analysis and facilitates metamorphic testing, where controlled variations in specifications produce predictable changes in test environments. By providing 
infrastructure for traceable, simulation-based testing, our work enables the SE community to 
validate complex, AI-driven systems.}

\vspace{-1em}
\subsection{Threats to Validity}

\textbf{Internal.} Our iterative LLM-based constraint solver may not converge for all constraint sets, and the batched resolution process could introduce order dependencies affecting solution quality. Additionally, physics-based relaxation may occasionally violate previously satisfied constraints, though our evaluation shows this occurs infrequently.
\textbf{External.} 
We
focus on indoor scenes, limiting generalizability to outdoor or specialized domains (e.g., underwater, space environments). The dataset may not fully capture real-world requirement complexity, and our constraint patterns derive primarily from residential and commercial spaces, potentially limiting applicability to industrial
environments.

\vspace{-1em}
\section{Related Work}

\textbf{2D UI Code Generation.}
Automated UI code generation has emerged as an important SE research area \textred{and automate a labor-intensive and error-prone phase of the software development lifecycle}. GPT-4o exhibits omission and misarrangement when generating code directly~\cite{dcgen2024}. Decomposition approaches like DCGen~\cite{dcgen2024} (15\% improvement), UICopilot~\cite{uicopilot2024} (hierarchical generation), and DeclarUI~\cite{declarui2024} (96.8\% coverage) address these issues. \textred{These methods mirror the SE principle of modularity, aiming to improve not just accuracy but also the structural quality and maintainability of the generated code.} Benchmarks like Design2Code~\cite{design2code2024}, DesignBench~\cite{designbench2024}, and WebCode2M~\cite{webcode2m2024} enable evaluation. \textred{Such benchmarks are essential for rigorously evaluating and comparing different code generation techniques from an SE perspective, providing the community with standardized tools for quality assurance.} LayoutCoder~\cite{layoutcoder2024} improves BLEU
via layout information. \textred{This focus on structure is critical, as well-formed code is a cornerstone of maintainable and scalable frontend engineering.} Despite advances, these methods struggle with complex layouts and interactive behaviors.

\textbf{3D Software Generation.}
3D software expansion demands automated approaches \textred{to support critical SE tasks such as testing, debugging, and verification, especially in safety-critical domains like autonomous systems and robotics}. Early probabilistic methods~\cite{chang2015text,chang2017sceneseer,savva2017scenesuggest,jiang2018configurable,fu2017adaptive,ma2018language,zhang2022fast} suffered from limited diversity, \textred{hindering the discovery of edge cases, a primary goal of thorough software quality assurance}. Then, deep learning methods~\cite{wei2023lego,liu2023clip,zhao2024roomdesigner,ye2024maan,lin2024instructscene,zhou2024gala3d,zhai2025echoscene,tang2024diffuscene,zhai2024commonscenes,yang2024physcene} dominated. \textred{From an SE standpoint, these methods automate the creation of complex test assets (i.e., 3D software) that would otherwise require significant manual effort and cost.} These learn from 3D-FRONT~\cite{fu20213d}, \textred{enabling more flexible and specification-driven test case generation}. View-based methods~\cite{huang2018holistic,nie2020total3dunderstanding,yang2021indoor,chatterjee20243d,dai2024acdc} require physical input, \textred{making them unsuitable for the SE objective of proactively generating virtual test environments that do not exist exist}. Procedural generation~\cite{procthor,infinigen2023infinite} lacks flexibility\textred{, a key activity for ensuring system robustness}. LLM-based approaches~\cite{feng2024layoutgpt,yang2024holodeck,gao2024graphdreamer,fu2025anyhome,aguina2024open,ccelen2024design,ocal2024sceneteller,yang2024llplace} leverage language models. \textred{This represents a shift towards specification-driven development, where high-level, human-centric requirements are translated into concrete software artifacts.} Holodeck~\cite{yang2024holodeck} uses GPT-4 but inherits scene graph limitations. Fundamental limitations in controllability, expressiveness, and verifiability impede SE adoption. \textred{Specifically, the lack of verifiability remains a major barrier for their use in rigorous engineering workflows.}
\section{Conclusion}

In this paper, we presented \scenethesistool, a novel approach to 3D software synthesis that decomposes the problem into four verifiable stages connected by \scenedsl, a formal IR. Our evaluation demonstrates that \scenethesistool achieves over 80\% requirement capture accuracy, satisfies more than 90\% of constraints, and improves visual quality by 42.8\% over state-of-the-art methods. By applying SE principles to 3D software generation, we enable the fine-grained control, verifiability, and maintainability required for practical deployment in safety-critical domains.
\section*{Acknowledgement}
We would like to thank the anonymous reviewers for their helpful and constructive feedback. 
This work was supported by the Research Grants Council of the Hong Kong Special Administrative Region, China (RGC Ref. No. SRFS2425-4S03 of the Senior Research Fellow Scheme).

\bibliographystyle{ACM-Reference-Format}
\bibliography{sample-base}

\end{document}